\documentclass[conference]{IEEEtran}
\IEEEoverridecommandlockouts
\usepackage{cite}
\usepackage{amsmath,amssymb,amsfonts}
\usepackage{algorithmic}
\usepackage{graphicx}
\usepackage{textcomp}
\usepackage{lipsum}
\usepackage{hyperref}
\usepackage{subcaption}
\usepackage{xcolor}
\usepackage{booktabs}
\usepackage{multirow}
\usepackage{float}
\usepackage{xspace}
\def\BibTeX{{\rm B\kern-.05em{\sc i\kern-.025em b}\kern-.08em
    T\kern-.1667em\lower.7ex\hbox{E}\kern-.125emX}}
\newcommand{\nxd}{P1\xspace} 
\newcommand{\cvc}{P2\xspace}
\newcommand{\nxe}{P3\xspace}

\begin{document}

\title{Soft-Sensing ConFormer: A Curriculum Learning-based Convolutional Transformer
}



\author{
\IEEEauthorblockN{Jaswanth Yella$^\dagger$}
\IEEEauthorblockA{Seagate Technology, MN, USA\\ \thanks{$\dagger$ University of Cincinnati, OH, USA}
Email: jaswanth.k.yella@seagate.com}\\
\IEEEauthorblockN{Yu Huang$^\mathparagraph$}
\IEEEauthorblockA{Seagate Technology, MN, USA\thanks{$\mathparagraph$ Florida Atlantic University, FL, USA}\\
Email:yu.1.huang@seagate.com}
\and
\IEEEauthorblockN{Chao Zhang$^\mathsection$}
\IEEEauthorblockA{Seagate Technology, MN, USA\\ \thanks{$\mathsection$ University of Chicago, IL, USA}
Email: chao.1.zhang@seagate.com}\\
\IEEEauthorblockN{Xiaoye Qian$^\$$}
\IEEEauthorblockA{Seagate Technology, MN, US\thanks{$\$$ Case Western Reserve University, OH, US}\\
    xiaoye.qian@seagate.com}\\
\IEEEauthorblockN{Sthitie Bom}
\IEEEauthorblockA{Seagate Technology, MN, USA\\
Email: sthitie.e.bom@seagate.com}
\and
\IEEEauthorblockN{Sergei Petrov$^{\ast}$}
\IEEEauthorblockA{Seagate Technology, MN, USA \thanks{$\ast$ Stanford University, CA, USA}\\
Email:sergei.petrov@seagate.com}\\
\IEEEauthorblockN{Ali A. Minai}
\IEEEauthorblockA{University of Cincinnati, OH, USA\\
Email: ali.minai@uc.edu}
}

\IEEEoverridecommandlockouts
\IEEEpubid{\makebox[\columnwidth]{This paper has been accepted by 2021 IEEE International Conference on Big Data \hfill} \hspace{\columnsep}\makebox[\columnwidth]{ }}

\maketitle

\begin{abstract}
Over the last few decades, modern industrial processes have investigated several cost-effective methodologies to improve the productivity and yield of semiconductor manufacturing. 
While playing an essential role in facilitating real-time monitoring and control, the data-driven soft-sensors in industries have provided a competitive edge when augmented with deep learning approaches for wafer fault-diagnostics. Despite the success of deep learning methods across various domains, they tend to suffer from bad performance on multi-variate soft-sensing data domains. To mitigate this, we propose a soft-sensing ConFormer (CONvolutional transFORMER) for wafer fault-diagnostic classification task which primarily consists of multi-head convolution modules that reap the benefits of fast and light-weight operations of convolutions, and also the ability to learn the robust representations through multi-head design alike transformers. 
Another key issue is that traditional learning paradigms tend to suffer from low performance on noisy and highly-imbalanced soft-sensing data. To address this, we augment our soft-sensing ConFormer model with a curriculum learning-based loss function, which effectively learns easy samples in the early phase of training and difficult ones later.
To further demonstrate the utility of our proposed architecture, we performed extensive experiments on various toolsets of Seagate Technology's wafer manufacturing process which are shared openly along with this work. To the best of our knowledge, this is the first time that curriculum learning-based soft-sensing ConFormer architecture has been proposed for soft-sensing data and our results show strong promise for future use in soft-sensing research domain.

\end{abstract}


\section{Introduction}

\begin{figure}[!t]
\centering
\includegraphics[width=3.5in]{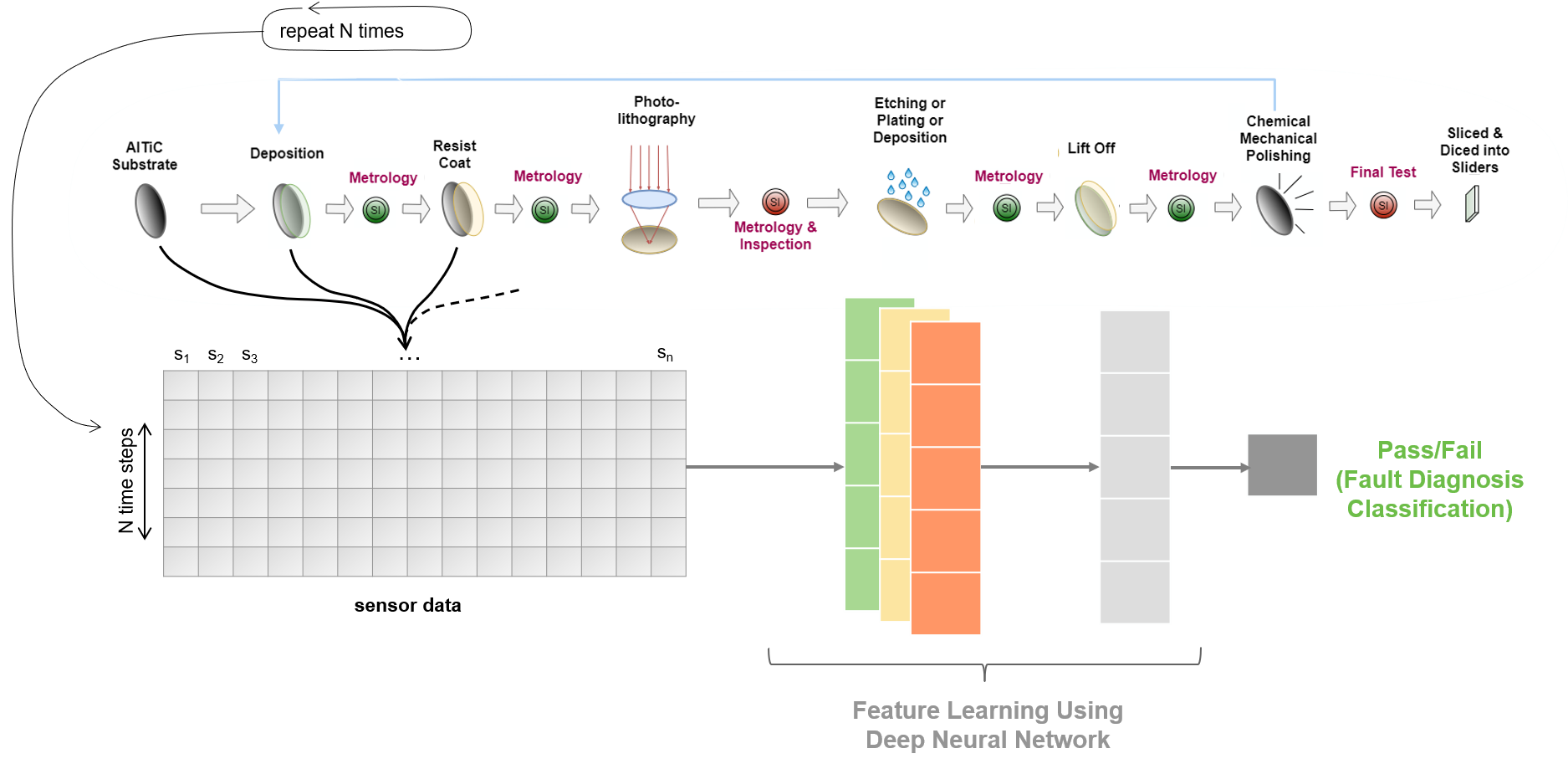}
\caption{An overview illustration of soft-sensing data processes and deep learning utility in fault-diagnosis at Seagate.}
\label{fig_overview}
\vspace{-1.5em}
\end{figure}

\textbf{Context and motivation.} With the increase in demand and competition in global manufacturing industries, a growing concern on production factors such as quality, safety, sustainability, and productivity have become prevalent over the last decade. This demands the modern industrial processes to turn towards crucial instruments for monitoring and control of process variables to react and act accordingly in real-time. However, such key process variables have high complex structure between them and measuring them online has economical and technical limitations. In recent years, soft-sensing technology has become an invaluable tool for online analysis and estimation of quality variables, including - but not limited to many industries such as chemical plants, nuclear power plants, pollution monitoring, and semi-conductor manufacturing industries \cite{savytskyi-softsensing}. These soft-sensors are defined as the combination of hardware and software models, where hard-to-measure variables are estimated based on other available process  variables and parameters. The large influx of sensor data from each sensor enables engineers further to analyze and identify variations in the wafer processing. However, the data acquired from soft-sensors are high-dimensional, temporal, redundant, imbalanced, and prone to outliers. A vast number of such issues could be attributed to transient and intermittent failures. Hence, at times distributional shifts are observed between training and test data splits resulting in drastic differences in model performance.  Therefore, the study of multi- variate time series wafer fault detection in soft-sensing technology is of great significance.

In the past, several machine learning based feature extraction methods have been proposed. Specifically, linear feature extractors such as Principal Component Analysis (PCA) \cite{jiang2015performance,yuan2016semisupervised}, Canonical Correlation Analysis (CCA) \cite{chen2016canonical}, and Partial Least Squares (PLS) \cite{yuan2018multi} have been utilized for fault-diagnosis problems. However, due to the inherently complex nature of soft-sensing data, these linear methods often fail to extract relevant features for classification or regression tasks in fault-diagnosis. To address this issue, deep learning methods have been adopted to extract and analyse unusual phenomena in the data. Recently, autoencoder variants such as the gated-stacked autoencoder \cite{sun2020deep} and quality stacked autoencoder \cite{yuan2019hierarchical} were proposed to overcome the weaknesses of shallow learning approaches. Recurrent neural network variants have had success in text and time-series domains, specifically, Long Short Term Memory (LSTM) networks have been utilized for modeling high-dimensional multivariate data \cite{ke2017soft, karim2017lstm}. However, these models fail to scale on soft-sensing data where temporal data shifts and frequent outliers are observed \cite{sun2021survey}. While transformers employs multi-head self attention blocks to alleviate the long-term dependency issues by learning global positional information \cite{vaswani2017attention}, they still require large amount of data and suffer from quadratic complexity. Convolutional neural networks, on the other hand, are fast, light-weight, and are adept at exploiting local information, but they still struggle to learn global context information efficiently \cite{han2020contextnet,gulati2020conformer,liu2021convtransformer}, which is essential for data with intrinsic data shifts. Apart from these nuances, the traditional learning paradigm of the deep learning models hinder the performance of the soft-sensing fault detection task due to the inherent noise and high-imbalance issues associated in the toolsets. To address the complexities of the soft-sensing toolset,  there is a need for an efficient model along with sophisticated learning strategy which sorts the training examples by difficulty and trains accordingly. 

\textbf{Our Contribution.} To address the aforementioned problems, we propose a curriculum learning-based soft-sensing convolutional transformer (ConFormer) which addresses three important issues. First, to efficiently exploit locality and low-level information in the multi-variate data, we utilize the simple, fast and light-weight 1-D convolution based modules in our architecture. Second, inspired by the multi-head attention design in the transformer, we define multi-head convolutional modules for learning global contextual information and pass these representations to a global average pooling layer which is processed further by a fully-connected layer for fault-diagnosis classification. And third, to deal with complex yet noisy soft-sensing data, we utilize a curriculum-learning based loss function i.e., SuperLoss \cite{castells2020superloss}, which assigns high confidence to less complex/easier samples and low confidence to difficult samples. To demonstrate the effectiveness of our proposed model, we apply it to multiple toolsets of Seagate Technology's real-time soft-sensing based wafer manufacturing toolsets and compare it with existing baselines. Our results show that the curriculum learning-based soft-sensing ConFormer outperforms existing baselines, demonstrating its utility. To further encourage novel solutions for the soft-sensing research domain, we open-source the toolset for public use. We provide a brief overview of Seagate's process management and machine learning utility for fault-diagnosis in figure \ref{fig_overview}.   With these aspects in mind, the novel contributions of the work are as follows: 
\begin{enumerate}
    \item We propose ConFormer, which replaces attention in the transformer architecture with multi-head 1-D convolutions to better capture representations in highly imbalanced historical soft-sensing data. We further augment the training procedure with SuperLoss, a curriculum learning-based loss function to train efficiently easier samples first and difficult samples later. To the best of our knowledge, this is the first time a curriculum learning-based soft-sensing ConFormer architecture has been proposed for soft-sensing multi-task classification.
    
    \item Experimental results compared with state-of-the-art models indicate that our proposed architecture demonstrates superior performance and utility for future use cases in soft-sensing domain.
    
    \item Our ablation studies suggest that curriculum learning using SuperLoss is helpful in improving performance for data that is heavily imbalanced and has many outliers. 
    
    \item To further encourage novel solutions in the future, we open-source Seagate Technology's wafer manufacturing tool-sets containing multi-variate time-series soft-sensing data specifically for fault diagnostics multi-task classification.
\end{enumerate}



\begin{figure*}[ht]
  \includegraphics[width=\textwidth,height=4cm]{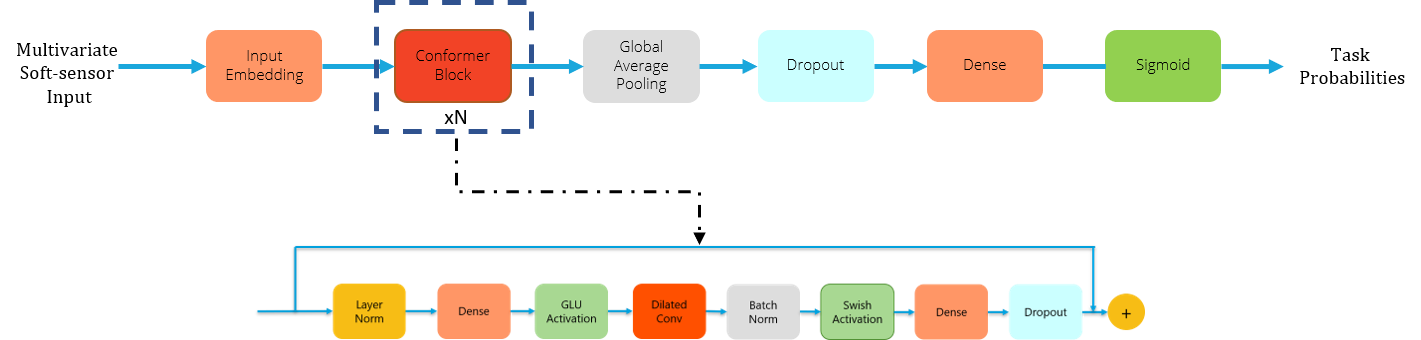}
  \caption{An overview of ConFormer Architecture. In the upper part of the figure, we show the conformer block which can be defined with multiple custom convolutional modules as shown in below figure. The convolutional block is a sandwich style declaration between two dense layers and a dilated convolution in the middle while regularizing the layers with batch normalization, dropout and swish activation functions. }
\end{figure*}
\section{Curriculum Learning-Based Soft-Sensing ConFormer}
In this section, we present the curriculum learning-based soft-sensing ConFormer. We introduce the soft-sensing conformer idea in section \ref{section-3.1}. In section \ref{section-3.2}, we briefly describe the curriculum learning paradigm and introduce the super-loss function utilized for this work.

\subsection{ConFormer: Convolutional Transformer}
\label{section-3.1}
In  multi-variate time-series data, a convolution is performed by applying a filter and sliding it over the series. In many recently proposed architectures, such convolution blocks are defined, applied and coupled with activation functions to achieve classification or regression. Recently, transformer designs have inspired many models where attention is replaced or coupled with convolutions. In this work, we achieve soft-sensing data classification through similar design and define the Convolutional TransFormer (ConFormer) architecture below.


\textbf{Input Embedding:} The initial input is provided to a dense layer initially which is defined as follows
$$ X = \sigma(FFN(input))$$
where $FFN(.)$ is a dense layer and $\sigma$ is sigmoid activation function. This way the input representation is reduced to a lower dimension space where $X \in R^D$ and $D$ is the dimension size less than soft-sensing input dimension. This embedded representation is passed to multiple ConFormer blocks. We define the components of a ConFormer block below.

\textbf{ConFormer Block:} The convolution module of the architecture essentially consists of a gated linear unit, a convolution layer and a residual skip connection layer.
First, the input $X$ is passed through a gated linear unit which is defined as
\begin{equation}
    X^{1}=W^{A} X \odot \operatorname{sigmoid}\left(W^{B} X\right)
\end{equation}
where $W^{A}$ and $W^{B}$ are trainable parameters associated with the gated linear unit layer. To further extract the abstract features from $X^{1}$, a 1-D dilated convolution \cite{yu2016multiscale} is applied to create feature maps with filters of kernel size $K$.
In our experiments, we used various kernel size for each convolution module. We utilize dilation due its effectiveness in skipping $d$ values of the input while improving the receptive field of input without additional pooling and with no loss of resolution. As a side-effect, this also allows the model to learn relations between data points that are far apart. The dilated convolution ($Conv(.)$) operation can be formulated as:
$$X^2 = Conv(X^1)$$

Then the convolved output is effectively regularized by applying batch normalization. We further utilize a self-gating activation function known as swish activation function \cite{ramachandran2017searching} which is given as $f(x) = x.\sigma(x)$.

Further, a feed forward network is applied along with dropout resulting in an output denoted as $X^{2'}$.
The output of the network $X^{2'}$ is summed with the original input $X$, thereby creating a skip connection in the module which helps in faster convergence and avoids overfitting. The output for $X^3$ is given as follows: 
\begin{equation}
    X^{3} = \operatorname{FFN}(X^{2'}) + X
\end{equation}
This sandwich style design of convolutional module is inspired by models proposed previously for natural language processing and speech recognition\cite{gulati2020conformer,wu2020lite}. 

\textbf{Global Average Pooling:} Upon receiving the outputs from multiple ConFormer blocks, we concatenate them and perform global averaging pool (GAP) operation as follows.
$$Z = GAP(Concat([X^{3}_{1}, X^{3}_{2}, X^{3}_{3}]))$$

where $GAP(.)$ is the global averaging pool function and $Concat(.)$ operation concatenates the output received from the 3 conformer blocks declared in the study. The global average pooling operation enables the model to learn global contextual information alike in Transformers\cite{vaswani2017attention}. Inspired by the CNNs global contextual learning capacity shown in \cite{hu2018squeeze}, we utilized squeeze-and-excitation inside the conformer block for learning global contextual information. However we found the simple $GAP$ operation over multiple ConFormer blocks to be much effective than utilizing squeeze-and-excitation within each ConFormer block. 

\textbf{Classification:} After performing GAP on the concatenated outputs of ConFormer blocks, we regularize the output $Z$ using dropout and apply another $FFN(.)$ along with sigmoid activation function for classifying the tasks. We denote them as follows:
$$Y = \sigma(FFN(\textit{Dropout}(Z)))$$

where $Dropout(.)$ function is applied to $Z$, then $FFN(.)$ is applied to convert the $Z$ into output dimension space, and then  sigmoid activation function ($\sigma$) is applied to convert the non-linear values of the layer to probability outputs. In our results, we show the effectiveness of the model on various toolsets of Seagate's wafer manufacturing plants.

\subsection{Super Loss: A Curriculum Learning-Based Loss Function}
\label{section-3.2}
While convolutional neural nets have evolved over the last few years to accommodate more layers, to reduce the size of the filters, and even to eliminate the fully-connected layers, relatively less attention has been paid to improving the training process. As a result, the traditional training procedure hinders the performance of the model due to the various types of noise (i.e., label noise and feature noise) present in the data \cite{zhu2004class}. Soft-sensing data invariably suffers from such noise due to corrupted/damaged sensors, intermittent/transient failures, or, label assignment consensus issues across different engineering teams. Moreover, heavy class/task imbalance in the soft-sensing data makes the performance of the model much worse. In this section, we address the soft-sensing data imbalance and noise issue by incorporating a robust loss function based on curriculum learning which organizes the samples based on their complexity and further exploits clean data efficiently.  

We first define the weighted binary cross-entropy loss for multi-task classification as:

\begin{equation}
\label{eqn:wbce}
\ell_{i}(y_i, \hat{y}_i) = -(\beta  y_i \log (\hat{y}_i)+(1-y_i) \log (1-\hat{y}_i))
\end{equation}
where $\beta$ is the support to deal with class/task imbalance, and $y_i$ and $\hat{y}_i$ are the ground truth and predictions of the sample, respectively. 
The support $\beta$ is given as $\beta = \frac{N}{2m*n_j^t}$ where $N$ is the total number of samples in the toolset, $m$ is the total number of tasks, and $n_j^t$ is the number of samples in the task.

In an ideal scenario, where the data is balanced and free from noise, the gradient updates for the clean labels are consistent during training, enabling the model to achieve stability and faster convergence. However, this is rare in soft-sensing data, hence the aforementioned weighted-binary cross entropy utilized heavily for class/task imbalance problems is not practical to learn robust representations and achieve ideal performance. To address this, we incorporate SuperLoss\cite{castells2020superloss} to dynamically formulate curriculum learning as:
\begin{equation}
\mathrm{L}_{\lambda}\left(\ell_{i}, \sigma_{i}\right)=\left(\ell_{i}-\tau\right) \sigma_{i}+\lambda\left(\log \sigma_{i}\right)^{2}
\end{equation}

where $\tau$ is a threshold to separate easy samples and hard samples based on an empirical value. In our experiments, we use $log(C)$, where $C$ is the number of classes/tasks involved in the tooslet. Here $\lambda$ is a regularizing parameter for the loss function which is set to 0.25 in our experiments, and $\sigma_{i}$ is the confidence associated with the loss which amplifies the contribution of easy samples and no improvement for hard samples. It can be solved in closed-form using an inverse function as:

\begin{equation}
\sigma_{\lambda}^{*}\left(\ell_{i}\right)=e^{-W\left(\frac{1}{2} \max \left(-\frac{2}{e}, \frac{\ell_{i}-\tau}{\lambda} \right)\right)}
\end{equation}

Here W stands for Lambert W function, which is an inverse function taking the the $l_i$ and $\tau$ as parameters to solve for the confidence value. The parameters $\sigma^*$ and $\tau$ are scalar units computed through back-propagation.


\section{Experimental Setup}
\begin{figure}[!t]
\centering
\includegraphics[width=2.5in]{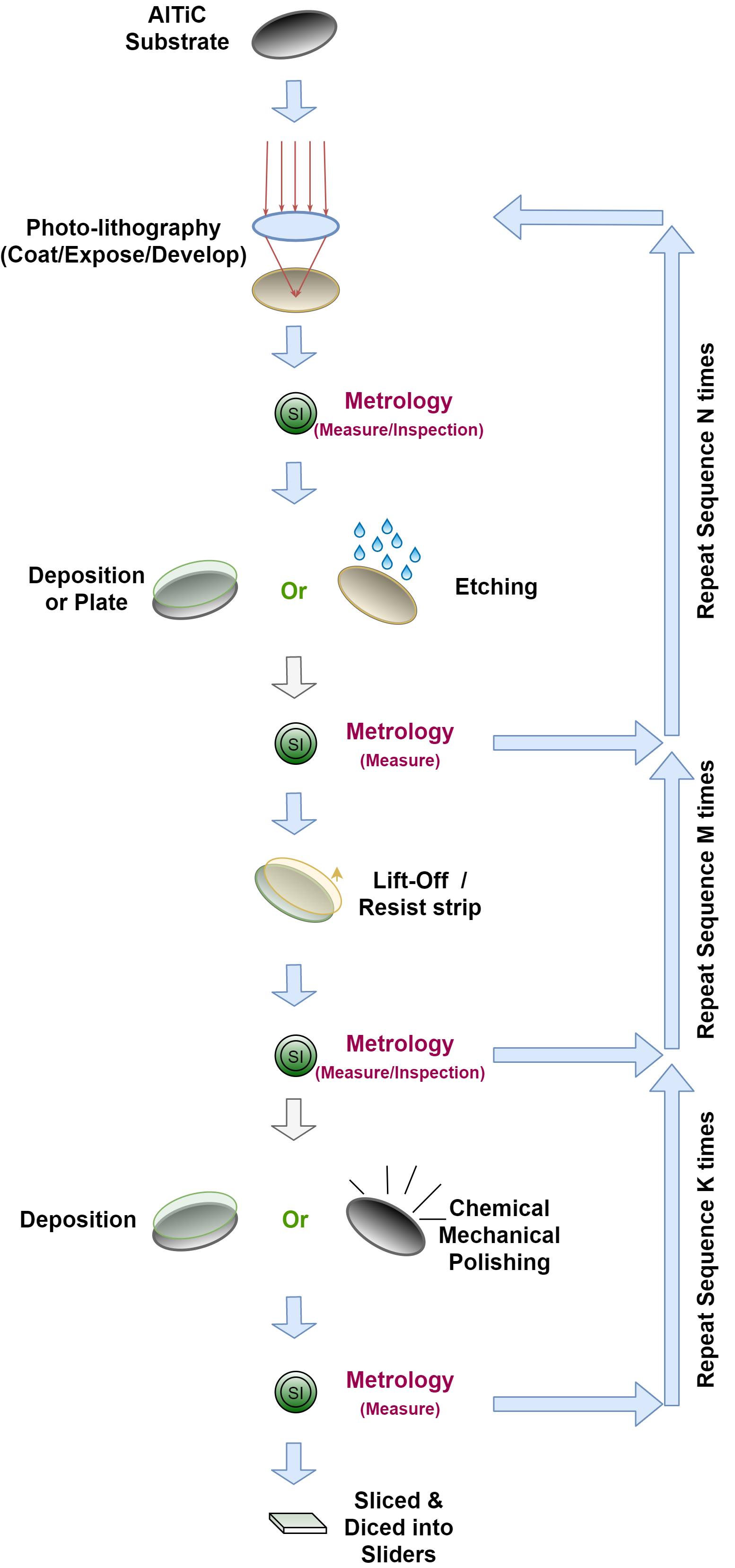}
\caption{An overview of wafer manufacturing process stages where the final measurement result determines the quality of the wafer.}
\label{fig_wafer}
\end{figure}

In this section, we provide details on the experiments conducted and compared with the baseline models to demonstrate the performance and efficiency of our curriculum learning based ConFormer. To perform these experiments, we use Seagate Technology's soft-sensing data  and evaluate on various toolsets that are openly available for comparison studies. The wafer manufacturing process is outlined in Fig. \ref{fig_overview} and \ref{fig_wafer} which are referred from \url{https://github.com/Seagate/softsensing_data}.

\subsection{Seagate Wafer Soft-sensing Tooslet} 

In a real-time industrial setting, the function mapping soft-sensor measurements inputs to an output measurement denoting the faults in the wafer as part of a binary classification task is extremely non-linear. This is due to the fact that measurements from various soft-sensors carry different information about characteristics and dynamics. Another factor is transient and intermittent faults that occur sporadically during the processes \cite{bakshi2014intermittent,satnamintermittent}. These faults are arbitrary and it is difficult to pinpoint their causes due to their complex nature. Such faults do not occur often, and when they do, they tend to disappear mysteriously. Complex toolsets with such faults are rare and extremely valuable -- especially in the soft-sensing domain -- to study the limitations and foster novel solutions. 

Seagate attempts to provide large scale soft-sensing toolsets that are queried and processed from Seagate wafer manufacturing factories. These real-time multi-variate time-series toolsets are retrieved from manufacturing machines that are high-dimensional and extremely imbalanced. Each wafer used for Seagate hard drives undergoes several processing stages such as metal deposition, dielectric deposition, etching, electroplating, planarization and lithography \cite{wiki:Semiconductor_device_fabrication}. At every processing stage, hundreds of soft-sensors are deployed in the processing machines to monitor the health of the wafer. For every few seconds, these sensors collect measurements of the wafer which are stored in Seagate's big data servers for downstream control and monitoring tasks.

\begin{figure}[!t]
\centering
\includegraphics[width=3in]{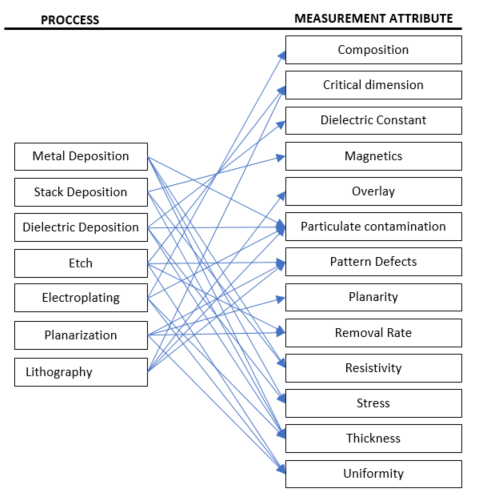}
\caption{Overview of the main categories of processes on the left and the corresponding critical measurement variables per each category on the right}
\label{fig_meas}
\end{figure}

Each wafer in process goes through several processing stages and is represented as a single time-series data point. At each processing stage, multiple critical measurements are estimated for each category as shown in Fig.\ref{fig_meas}. These estimates of each category are continuous values calculated through non-linear methods. The engineers at Seagate inspect these measurements and attest to the quality of wafer based on some internal heuristic threshold values for each category. For the sake of simplicity, we retrieved the discrete information of wafer diagnosis i.e., pass or fail, as suggested by the Seagate's wafer manufacturing team. This simplifies the classification task by mapping the time-series measurements from each processing stage of the wafer into a multi-task binary classification problem. 

\subsection{Pre-processing}

\begin{table}[!t]
\renewcommand{\arraystretch}{1.3}
\caption{Summary for the data sets}
\label{tbl:data}
\centering
\begin{tabular}{rrrrrrr}
\hline
 & \multicolumn{2}{c}{\nxd} & \multicolumn{2}{c}{\nxe} & \multicolumn{2}{c}{\cvc} \\
Task &pos&neg&pos&neg&pos&neg\\
\hline
1&295&8328&256&6433&109&2496 \\
2&40&12747&773&26811&335&12857 \\
3&291&56198&2069&78844&46&1026 \\
4&188&14697&582&27809&15&4180 \\
5&568&40644&247&9652&300&22254 \\
6&863&84963&884&27337&166&40811 \\
7&2501&153970&2108&53921&875&75706 \\
8&490&2919&2016&77473&1097&18890 \\
9&104&29551&644&23305&537&4247 \\
10&57&10813&270&25651&1547&129914 \\
11&306&47219&3792&354328&& \\
\hline
\end{tabular}
\end{table}

In total, we provide 3 toolsets of soft-sensing data where each toolset covers 92 weeks of data. The data is split as follows: first 70 weeks as the training set, the following 14 weeks as the validation set, and the remaining 8 weeks as the test set. The raw data of each tooslet is further pre-processed for use with machine learning models. This includes removing redundant sequences, imputing missing values, and scaling the features using a min-max scaler. We further include necessary auxillary information by concatenating process-relevant categorical variables to the time-series data. However, to preserve confidentiality, we de-identify the toolsets, which includes anonymizing the data headers. In the following sub-sections, we provide more details on each toolset.

The toolsets are as follows:

\textbf{\nxd:} In the \nxd toolset, there are 90 sensors capturing information related to deposition and etching steps. The data is captured from these sensors for every second and is pre-processed as mentioned earlier. The post-processed data is split into training, validation, and test sets. There are 194k samples in the training set, 34k samples in the validation set, and 27k in the test set. Each toolset has a maximum of 2 timesteps and we
pad the instances that are less than the maximum length. We further concatenate the one-hot encoded categorical variables with the sensor data, thus yielding a total of 817 features which includes the 90 sensors data values as features. There are 11 measurement tasks represented as binary classes/tasks with about 1.2\% of them as positive samples. 

\textbf{\cvc:} The pre-processing for the \cvc toolset is carried out similar to \nxd. The maximum timseries length is 2. Here 2 indicates, 2 different measurements of the same wafer at different time points.
The data consists of 498 features where 43 of them are sensor measurements and the rest are categorical variables associated with \cvc toolset. In total, the data contains 205k training, 35k  validation, and 20k testing samples. There are 10 measurement tasks identified as binary classes/tasks with 1.6\% of them are positive samples.

\textbf{\nxe:} The \nxe tooslet is just as heavily imbalanced as \nxd and \cvc. There are 57 sensors and in total there are 1484 features with categorical variables. In this toolset, there are 457k samples for training, 80k for validation, and 66k are test. There are 11 measurements to be classified similar to the other toolsets and 1.9\% of the samples are positive.

The number of positives and negatives of each task for all the toolsets are shown in table \ref{tbl:data}. All the data for the aforementioned toolsets is available publicly at https://github.com/Seagate/softsensing data.

\subsection{Baseline Methods}
To demonstrate the superiority of our model, we compare it with two baseline models. To ensure fairness, during our evaluation we used the same toolset splits i.e., training, validation, and test sets across all the methods.

The following baseline methods are used:

\textbf{LSTM:} Long Short Term Memory (LSTM) is a special variant of Recurrent Neural Networks (RNN) introduced by Hochreiter and Schmidhuber \cite{hochreiter1997long}. A standard RNN suffers from the vanishing gradient issue. In order to tackle that issue, LSTMs incorporate gating functions into their state dynamics. In the past, Ke et al. \cite{ke2017soft} have applied LSTMs on soft-sensing data for regression tasks. In this work, we use them as baseline for multi-task classification and use the weighted binary cross-entropy loss to handle the imbalance issue.




\textbf{MLSTM-FCN:} In this work, Karim et al. propose an architecture that combines LSTM and CNN models for multi-variate time-series classification tasks \cite{karim2019multivariate}. They utilize squeeze and excite blocks \cite{hu2018squeeze} for 1-D convolutions that adaptively recalibrate the input feature maps by exploiting the contextual information outside the local receptive field. 


\subsection{Training and Hyper-parameter Setting}
We implemented our model in Keras 2.3. We tested our model on multiple hyper-parameters and chose the best performing values. For embedding size, we chose 64 after outperforming from the pool of size parameters [32,64,128,256]. Similarly, we tried different dropout values from the set ranging from no dropout to 0.8 dropout and chose 0.5. We further regularized the model by trying different values in the set [1e-3,1e-4,1e-5,1e-6] and chose 1e-4 as our regularizing value. For optimization, we used adam \cite{kingma2017adam} with the scheduled learning rate ($lr$) set as follows:

\begin{equation}
\operatorname{lr}=0.1 * d^{-0.5} * \min \left(\text { step }^{-0.5}, \text { step } * \operatorname{warmup}^{-1.5}\right)
\end{equation}
Here $step$ refers to the epoch step, $d$ is dimension of the embedding vector, and $warmup$ is set to 4000. We trained our model for 500 epochs along with early-stopping based on loss with patience set to 100 epochs. The experiments for this work were trained on an AWS p3.2xlarge instance with 16 GB NVIDIA Tesla V100 GPU.

\section{Results}

In this section, we compare our proposed curriculum learning-based ConFormer with the existing state-of-the-art models proposed for multivariate time-series classification tasks. To evaluate and compare the performance of the models, we report Area Under Receiver Operating Curve (AUROC) scores for each toolset used in the experiments. During our experiments, instead of individually tuning models for each toolset, we aimed to design a single generalizable model for all the toolsets. This is a critical aspect of this work, as one of the primary challenges is to find optimal hyper-parameters for each model, which is not practical for a data with temporal data shifts and noise. 

The curriculum learning-based soft sensing ConFormer shows promising results without the need for excessive parameter tuning. In previous work, Gulati et al \cite{gulati2020conformer} demonstrated the utility of swish activation function and suggested its benefit in convergence of deep learning models. We attest to this finding and further suggest that the swish activation also eliminates the need for identifying the regularizing parameters for the kernel weights of convolutional and dense network modules, while providing optimal performance. Indeed, the swish activation function also enabled the model to use a dropout of only 0.15 which suggests that ConFormer architecture is effectively regularizing the layers without the need for providing external regularization of the weights, whereas in the case of other architectures i.e., MLSTM-FCN and LSTM models, a significant amount of time has to be spent on finding the best possible parameters for the kernel regularizer weights and dropout to handle the imbalance issue. TO further demonstrate the quality and performance of the model, in the following sections we will review the results obtained on the tool sets of Seagate.

\textbf{Results on \nxd:} In table \ref{tbl:nxd}, we provide the comparison results for the \nxd toolset. Here we observe that ConFormer model outperforms the baseline models in most of the tasks highlighted in the table. Although, the imbalance ratio of the \nxd toolset is 1.2\%, our proposed model is unaffected by the ratio, whereas in other models which performed <0.8 AUROC score certainly seem to be affected by the unavailability of positives. Some of the lower task performance of the baselines is also due to the hidden outliers in the toolset. Our model achieves at least a 20 percent improvement on 11 soft-sensing measurements task classification for \nxd with 650k trainable parameters. The improvement is much greater on some tasks.

\begin{table}[!t]
\renewcommand{\arraystretch}{1.3}
\caption{An overall AUROC performance comparision of methods on \nxd toolset}
\label{tbl:nxd}
\centering
\begin{tabular}{llll}
\toprule
       Task &                ConFormer &                MLSTM-FCN &                LSTM \\
\midrule
Task-1 &  \textbf{0.716}$\pm$0.077 &   0.68$\pm$0.115 &   0.681$\pm$0.08 \\
  Task-2 &  0.716$\pm$0.075 &   0.609$\pm$0.07 &   \textbf{0.74}$\pm$0.086 \\
  Task-3 &  0.806$\pm$0.004 &   0.82$\pm$0.041 &  \textbf{0.823}$\pm$0.007 \\
  Task-4 &  \textbf{0.884}$\pm$0.017 &  0.874$\pm$0.014 &  0.871$\pm$0.009 \\
  Task-5 &  \textbf{0.665}$\pm$0.028 &   0.547$\pm$0.01 &  0.521$\pm$0.037 \\
  Task-6 &  0.637$\pm$0.041 &  \textbf{0.667}$\pm$0.033 &  0.484$\pm$0.027 \\
  Task-7 &  0.653$\pm$0.011 &  0.652$\pm$0.027 &  \textbf{0.667}$\pm$0.006 \\
  Task-8 &  0.584$\pm$0.005 &    \textbf{0.8}$\pm$0.033 &  0.676$\pm$0.106 \\
  Task-9 &  0.754$\pm$0.061 &  \textbf{0.799}$\pm$0.046 &  0.683$\pm$0.092 \\
 Task-10 &  \textbf{0.944}$\pm$0.004 &  0.868$\pm$0.052 &   0.841$\pm$0.09 \\
 Task-11 &  \textbf{0.878}$\pm$0.021 &  0.748$\pm$0.084 &  0.799$\pm$0.012 \\
 
\bottomrule

\end{tabular}
\end{table}
\textbf{Results on \cvc:} Similar to the \nxd experiment, we test the models on the \cvc toolset. Unlike \nxd, this toolset has only 10 measurements for classification. However, the imbalance ratio is still similar to the other toolsets. However, since we're splitting our toolsets based on time, some of the tasks have do not have any positives or negatives available for that time-frame. Hence, one may observe no scores to be reported for those tasks. Our results in table \ref{tbl:cvc} suggest that all the models equally suffer from sparsity of the data. Although ConFormer performance superior in Task-2 and Task-5, the MLSTM-FCN model performance is better than ConFormer in Task-3, Task-7, Task-8 and Task-9. However, with our constraint of an auroc score over 0.8 for consideration of deployment, the superiority of the MLSTM-FCN performance is still not considered to be deployable due to it's sub-par efficiency. 

\begin{table}[!t]
\renewcommand{\arraystretch}{1.3}
\caption{An overall AUROC performance comparision of methods on \cvc toolset}
\label{tbl:cvc}
\centering
\begin{tabular}{llll}
\toprule
       Task &                ConFormer &                MLSTM-FCN &                LSTM \\
\midrule
Task-1 &   0.28$\pm$0.049 &   0.29$\pm$0.077 &   \textbf{0.35}$\pm$0.094 \\
  Task-2 &  \textbf{0.936}$\pm$0.012 &  0.923$\pm$0.008 &  0.91$\pm$0.014 \\
  Task-3 &  0.495$\pm$0.063 &  \textbf{0.665}$\pm$0.084 &   0.56$\pm$0.137 \\
  Task-4 &      --$\pm$-- &      --$\pm$-- &      --$\pm$-- \\
  Task-5 &  \textbf{0.542}$\pm$0.055 &     0.54$\pm$0.1 &  0.512$\pm$0.012 \\
  Task-6 &      --$\pm$-- &      --$\pm$-- &      --$\pm$-- \\
  Task-7 &  0.50$\pm$0.033 &  \textbf{0.596}$\pm$0.018 &    0.591$\pm$0.032 \\
  Task-8 &    0.5$\pm$0.029 &  \textbf{0.702}$\pm$0.038 &   0.51$\pm$0.047 \\
  Task-9 &      --$\pm$-- &      --$\pm$-- &      --$\pm$-- \\
 Task-10 &  0.806$\pm$0.026 &  \textbf{0.831}$\pm$0.003 &  0.82$\pm$0.007 \\
\bottomrule

\end{tabular}
\end{table}

\textbf{Results on \nxe:} The \nxe toolset is comparatively larger than the other toolsets in terms of the number of features and training points involved. More details of the toolset are given in the previous section. Due to its high dimensionality, when compared to \nxd and \cvc, this toolset takes at least 50 seconds per epoch, whereas \nxd and \cvc average around 35 seconds per epoch. The model for \nxe takes at least 45 mins to 1 hour to converge to a solution on the system used. When compared to the baseline models, the ConFormer model achieve atleast 10\% improvement in classifying the soft-sensing measurements, and much more in some cases. Also, note that LSTM has performed equally better in certain tasks especially Task-7 which is greater than 0.8 auroc score. This suggests that LSTM model may act as complimentary for task-specific deployment.

\begin{table}[!t]
\renewcommand{\arraystretch}{1.3}
\caption{An overall AUROC performance comparision of methods on \nxe toolset}
\centering
\begin{tabular}{llll}
\toprule
       Task &                ConFormer &                MLSTM-FCN &                LSTM \\
\midrule
 Task-1 &   \textbf{0.874}$\pm$0.008 &   0.87$\pm$0.007 &  0.868$\pm$0.003 \\
  Task-2 &  \textbf{0.702}$\pm$0.006 &  0.642$\pm$0.011 &  0.702$\pm$0.003 \\
  Task-3 &  \textbf{0.685}$\pm$0.024 &   0.602$\pm$0.02 &  0.619$\pm$0.024 \\
  Task-4 &  \textbf{0.855}$\pm$0.014 &  0.754$\pm$0.009 &  0.825$\pm$0.011 \\
  Task-5 &  0.446$\pm$0.037 &  0.461$\pm$0.021 &  \textbf{0.535}$\pm$0.046 \\
  Task-6 &  0.712$\pm$0.035 &  0.694$\pm$0.032 &  \textbf{0.725}$\pm$0.009 \\
  Task-7 &  0.796$\pm$0.019 &  0.808$\pm$0.007 &  \textbf{0.813}$\pm$0.001 \\
  Task-8 &  0.759$\pm$0.008 &  0.688$\pm$0.003 &  \textbf{0.772}$\pm$0.005 \\
  Task-9 &  0.604$\pm$0.088 &  0.418$\pm$0.133 &  \textbf{0.773}$\pm$0.051 \\
 Task-10 &  0.762$\pm$0.021 &  0.719$\pm$0.016 &   \textbf{0.78}$\pm$0.016 \\
 Task-11 &  \textbf{0.837}$\pm$0.005 &  0.836$\pm$0.005 &  0.80$\pm$0.004 \\
\bottomrule

\end{tabular}
\end{table}
\begin{figure}[ht]
\centering
\includegraphics[width=3in]{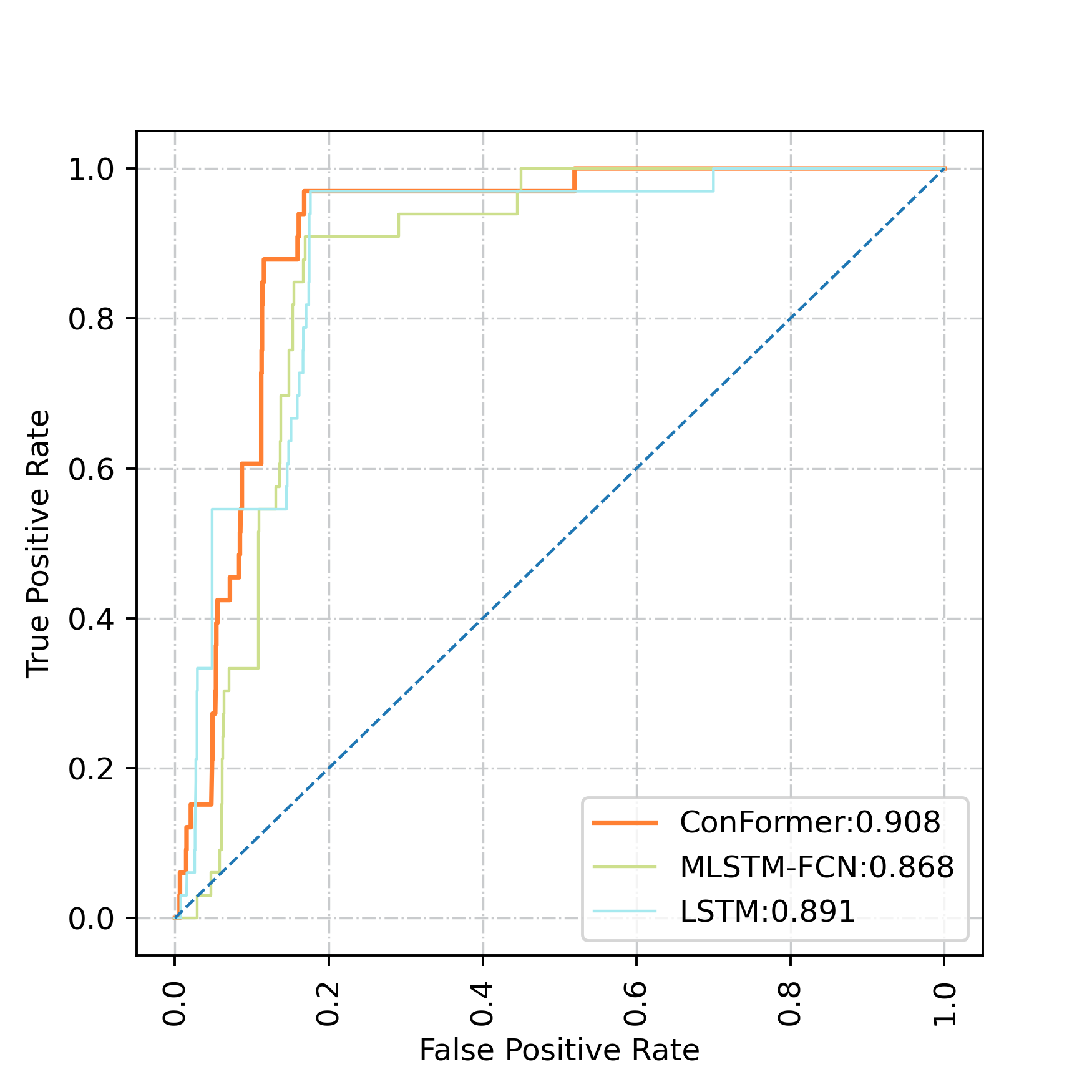}
\caption{An AUROC plot with ConFormer and its baseline models using \nxd toolset on Task-4}
\label{fig:nxdfig1}
\end{figure}

\begin{figure}[!t]
  \centering
  \includegraphics[width=3in]{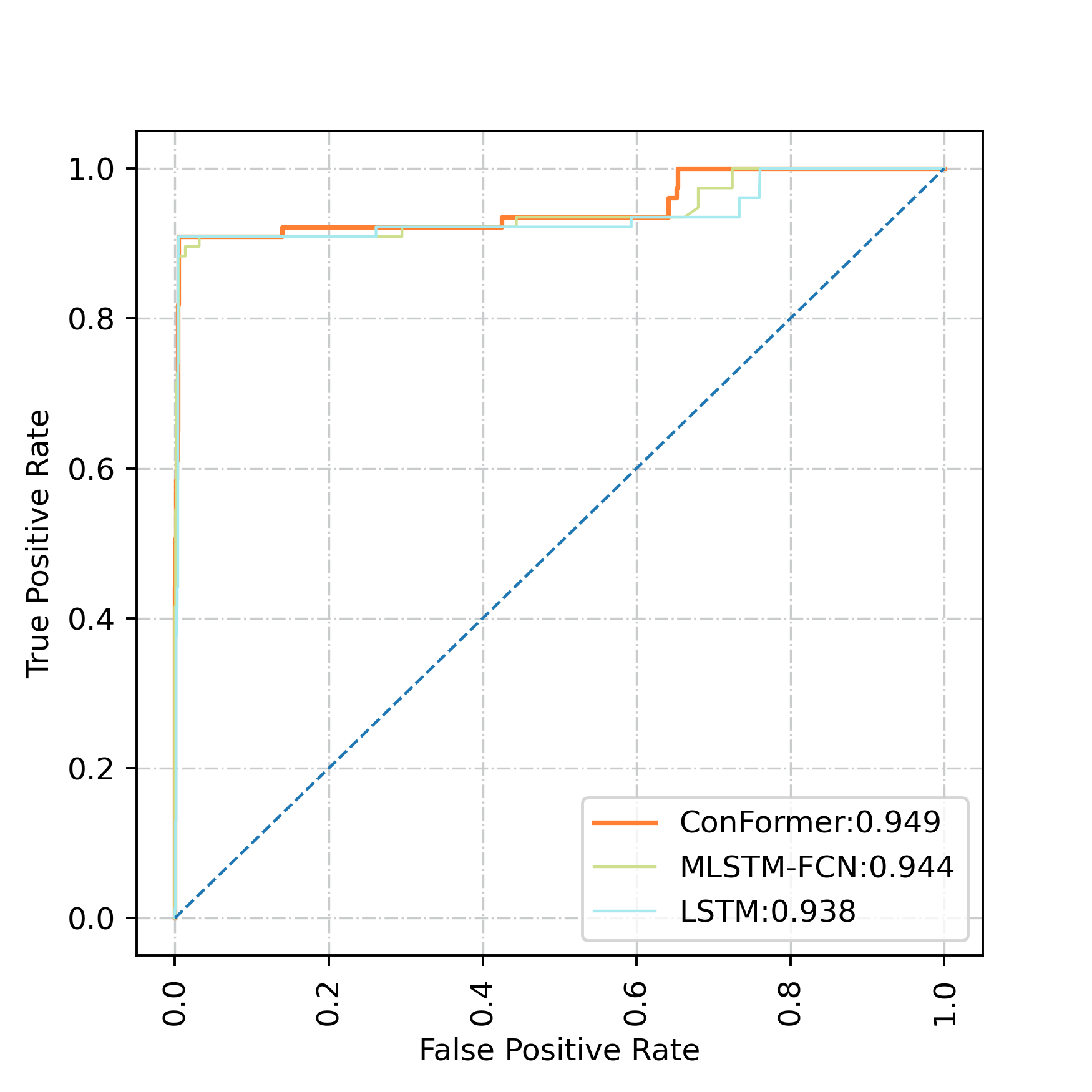}
  \caption{An AUROC plot with ConFormer and its baseline models using \cvc toolset on Task-2}
  \label{fig:nxdfig3}
\end{figure}

\begin{figure}[!t]
  \centering
  \includegraphics[width=3in]{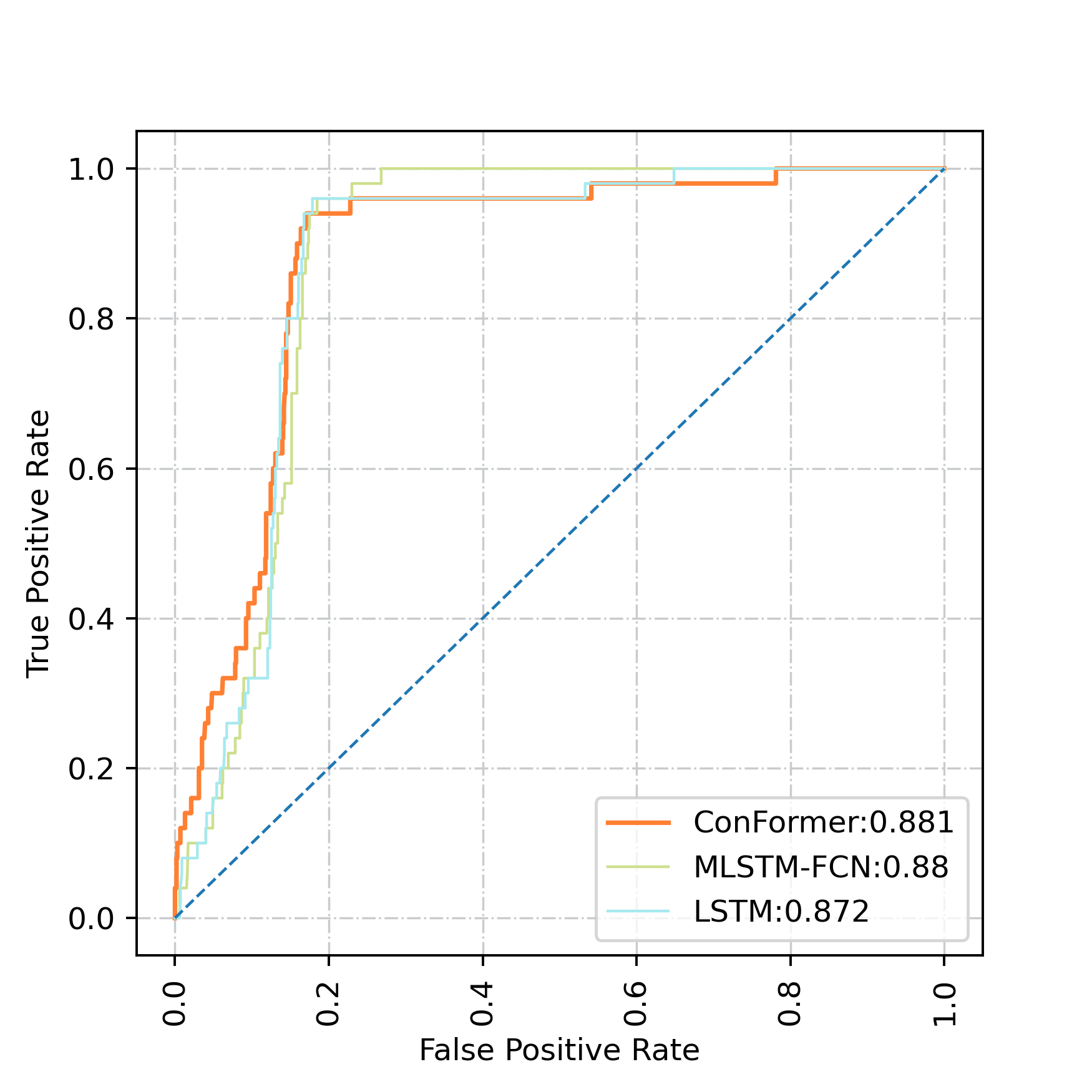}
  \caption{An AUROC plot with ConFormer and its baseline models using \nxe toolset on Task-1}
  \label{fig:nxdfig3}
\end{figure}

Overall, the studies on multiple toolsets suggest that our curriculum learning-based ConFormer model is effective in learning robust and helpful representations of soft-sensing data. In our experiments, we emphasized on generalized models and we didnot perform intensive hyper-parameter tuning for each specific toolsets due to the explosive number of parameters to be considered for each toolset. Although in certain tasks of toolsets (P1,P2 and P3), the ConFormers homogeneity couldn't be effeciently leveraged. One may take benefit from toolset based hyper-parameter tuning. An another direction could be an ensemble approach since one can observe the task performance of each model is complimentary
While transformers have shown to be effective in the NLP domain, our initial concern pertained to the hyper-parameter search and most importantly the quadratic time complexity of the model ($O(L.N^2)$ where $L$ is number of heads). The existing work on Convolutional Transformers \cite{gulati2020conformer,liu2021convtransformer}, show that transformers and ConFormers achieve similar performance but the ConFormer can perform with lower time complexity ($O(L.N)$). However, at the time of writing this paper, few alternate research investigations have been conducted utilizing transformers, graph neural networks, and other deep learning paradigms for soft-sensing data and we suggest readers to check these works \cite{zhang2021autoencoder, chao2021soft, xiaoye2021soft, yu2021grassnet}.

The results, when compared with baseline models, suggest that ConFormer model is able to leverage multiple contexts in learning representation i.e., local exploitation of convolutions and global information extraction through multi-head convolution global average pooling. At the same time, the curriculum learning-based SuperLoss function further augments the model performance via an easy-to-difficult learning paradigm. 

\subsection{Ablation Studies and Robustness Testing}
In order to test the effectiveness of our proposed architecture, in this section we show some further tests that we applied to the curriculum learning-based ConFormer for soft-sensing data. To simplify our experiments, we narrowed down our studies to Task-4 in P1 tool-set. However, we observed similar performance on other tasks and tool-sets.

\textbf{Benefit of Using Curriculum Learning}: To test the effectiveness of curriculum learning loss function i.e., SuperLoss, we performed a study where we used the weighted binary cross-entropy loss instead. In the first two rows of Table \ref{tbl:robust}, we show that curriculum learning based SuperLoss helps in consistently achieving better performance. The 0\% column indicate there is no corruption of the labels involved.

\textbf{Robustness Testing:} In this experiment, we corrupt the labels by arbitrarily flipping a percentage set of label class/task in the training data and study the robustness of our proposed model on the test set. The motivation behind this study is that soft-sensing labels tend to be classified as false positives or false negatives for a couple of reasons. First, although multiple teams in Seagate strictly scrutinize the quality of the wafer, due to consensus conflict which may arise across global engineering teams, a wafer could be mis-classified. Second, since we simplify the problem by converting the continuous value to a binary value based on an internal heuristic threshold value, there exist the possibility of inherent corruption associated with reliance on the threshold of the continuous value output received to distinguish the wafer as a pass or fail. We compare our model with existing models and show their performance in table \ref{tbl:robust} for Task-4 of \nxd toolset. 
We chose Task-4 of \nxd toolset to report the recall scores as it has one of the best performance among other tasks across all models despite the task sample imbalance for binary classification. Our recall scores for this test indicate strongly that the ConFormer model is robust to the noise when compared with other baseline models even without SuperLoss. Whereas other baseline models suffered with performance especially with corruption involved. This is predominantly due to the noise and imbalance associated in the toolset. A subtle change in the toolset is completely affecting the baseline models performance. However, utilizing a curriculum learning-based loss function in our framework improves robustness further. This suggests that our proposed architecture is able to learn helpful representations despite the strong imbalance in the tooslet.

\begin{table}[!t]
\renewcommand{\arraystretch}{1.3}
\caption{Robustness test performance of Task-4 with recall as assessment metric. Percentage in the header indicate the percentage of positive and negative labels corrupted in training set. }
\label{tbl:robust}
\centering
\begin{tabular}{lllll}
\toprule
 Model &       0\% &    20\% &       40\% &       60\% \\
\midrule
ConFormer-SuperLoss &  0.812 &  0.604 &  0.771 &  0.667 \\
ConFormer-BCE &  0.799 &  0.146 &  0.083 &  0.375 \\
MLSTM-FCN &  0.791 &  0.000 &  0.000 &  0.000 \\
LSTM &  0.812 &  0.000 &  0.000 &  0.000 \\
\bottomrule
\end{tabular}
\end{table}

\section{Conclusion}
In this work, we propose a curriculum learning-based ConFormer (CONvolutional transFORMER) for soft-sensing data classification. The soft-sensing data is extremely complex, imbalanced, and noisy. Due to these factors, traditional learning mechanisms and existing models fail to achieve satisfactory performance. We utilize a convolution-based architecture in a multi-head transformer design to improve on the limitations of existing convolutional models and augment it with a curriculum learning-based loss function. Our systematic experiments on various soft-sensing toolsets from Seagate manufacturing processes suggest promising utility for our proposed framework for use and future extensibility in the soft-sensing research domain.

\section*{Acknowledgment}
We sincerely thank Seagate Technology for the support on this study, the Seagate Lyve Cloud team for providing the data infrastructure, and the Seagate Open Source Program Office for open sourcing the data sets and the code. Special thanks to the Seagate Data Analytics and Reporting Systems team for inspiring the discussions.



\bibliographystyle{./bibliography/IEEEtran}
\bibliography{./bibliography/references}

\begin{thebibliography}{10}
\providecommand{\url}[1]{#1}
\csname url@samestyle\endcsname
\providecommand{\newblock}{\relax}
\providecommand{\bibinfo}[2]{#2}
\providecommand{\BIBentrySTDinterwordspacing}{\spaceskip=0pt\relax}
\providecommand{\BIBentryALTinterwordstretchfactor}{4}
\providecommand{\BIBentryALTinterwordspacing}{\spaceskip=\fontdimen2\font plus
\BIBentryALTinterwordstretchfactor\fontdimen3\font minus
  \fontdimen4\font\relax}
\providecommand{\BIBforeignlanguage}[2]{{%
\expandafter\ifx\csname l@#1\endcsname\relax
\typeout{** WARNING: IEEEtran.bst: No hyphenation pattern has been}%
\typeout{** loaded for the language `#1'. Using the pattern for}%
\typeout{** the default language instead.}%
\else
\language=\csname l@#1\endcsname
\fi
#2}}
\providecommand{\BIBdecl}{\relax}
\BIBdecl

\bibitem{savytskyi-softsensing}
O.~Savytskyi, M.~Tymoshenko, O.~Hramm, and S.~Romanov, ``Application of soft
  sensors in the automated process control of different industries,'' \emph{E3S
  Web of Conferences}, vol. 166, p. 05003, 01 2020.

\bibitem{jiang2015performance}
Q.~Jiang, X.~Yan, and B.~Huang, ``Performance-driven distributed pca process
  monitoring based on fault-relevant variable selection and bayesian
  inference,'' \emph{IEEE Transactions on Industrial Electronics}, vol.~63,
  no.~1, pp. 377--386, 2015.

\bibitem{yuan2016semisupervised}
X.~Yuan, Z.~Ge, B.~Huang, Z.~Song, and Y.~Wang, ``Semisupervised jitl framework
  for nonlinear industrial soft sensing based on locally semisupervised
  weighted pcr,'' \emph{IEEE Transactions on Industrial Informatics}, vol.~13,
  no.~2, pp. 532--541, 2016.

\bibitem{chen2016canonical}
Z.~Chen, S.~X. Ding, K.~Zhang, Z.~Li, and Z.~Hu, ``Canonical correlation
  analysis-based fault detection methods with application to alumina
  evaporation process,'' \emph{Control Engineering Practice}, vol.~46, pp.
  51--58, 2016.

\bibitem{yuan2018multi}
X.~Yuan, J.~Zhou, Y.~Wang, and C.~Yang, ``Multi-similarity measurement driven
  ensemble just-in-time learning for soft sensing of industrial processes,''
  \emph{Journal of Chemometrics}, vol.~32, no.~9, p. e3040, 2018.

\bibitem{sun2020deep}
Q.~Sun and Z.~Ge, ``Deep learning for industrial kpi prediction: When ensemble
  learning meets semi-supervised data,'' \emph{IEEE Transactions on Industrial
  Informatics}, vol.~17, no.~1, pp. 260--269, 2020.

\bibitem{yuan2019hierarchical}
X.~Yuan, J.~Zhou, B.~Huang, Y.~Wang, C.~Yang, and W.~Gui, ``Hierarchical
  quality-relevant feature representation for soft sensor modeling: a novel
  deep learning strategy,'' \emph{IEEE transactions on industrial informatics},
  vol.~16, no.~6, pp. 3721--3730, 2019.

\bibitem{ke2017soft}
W.~Ke, D.~Huang, F.~Yang, and Y.~Jiang, ``Soft sensor development and
  applications based on lstm in deep neural networks,'' in \emph{2017 IEEE
  Symposium Series on Computational Intelligence (SSCI)}.\hskip 1em plus 0.5em
  minus 0.4em\relax IEEE, 2017, pp. 1--6.

\bibitem{karim2017lstm}
F.~Karim, S.~Majumdar, H.~Darabi, and S.~Chen, ``Lstm fully convolutional
  networks for time series classification,'' \emph{IEEE access}, vol.~6, pp.
  1662--1669, 2017.

\bibitem{sun2021survey}
Q.~Sun and Z.~Ge, ``A survey on deep learning for data-driven soft sensors,''
  \emph{IEEE Transactions on Industrial Informatics}, 2021.

\bibitem{vaswani2017attention}
A.~Vaswani, N.~Shazeer, N.~Parmar, J.~Uszkoreit, L.~Jones, A.~N. Gomez,
  {\L}.~Kaiser, and I.~Polosukhin, ``Attention is all you need,'' in
  \emph{Advances in neural information processing systems}, 2017, pp.
  5998--6008.

\bibitem{han2020contextnet}
W.~Han, Z.~Zhang, Y.~Zhang, J.~Yu, C.-C. Chiu, J.~Qin, A.~Gulati, R.~Pang, and
  Y.~Wu, ``Contextnet: Improving convolutional neural networks for automatic
  speech recognition with global context,'' \emph{arXiv preprint
  arXiv:2005.03191}, 2020.

\bibitem{gulati2020conformer}
A.~Gulati, J.~Qin, C.-C. Chiu, N.~Parmar, Y.~Zhang, J.~Yu, W.~Han, S.~Wang,
  Z.~Zhang, Y.~Wu, and R.~Pang, ``Conformer: Convolution-augmented transformer
  for speech recognition,'' 2020.

\bibitem{liu2021convtransformer}
Z.~Liu, S.~Luo, W.~Li, J.~Lu, Y.~Wu, S.~Sun, C.~Li, and L.~Yang,
  ``Convtransformer: A convolutional transformer network for video frame
  synthesis,'' 2021.

\bibitem{castells2020superloss}
T.~Castells, P.~Weinzaepfel, and J.~Revaud, ``Superloss: A generic loss for
  robust curriculum learning,'' \emph{Advances in Neural Information Processing
  Systems}, vol.~33, 2020.

\bibitem{yu2016multiscale}
F.~Yu and V.~Koltun, ``Multi-scale context aggregation by dilated
  convolutions,'' 2016.

\bibitem{ramachandran2017searching}
P.~Ramachandran, B.~Zoph, and Q.~V. Le, ``Searching for activation functions,''
  2017.

\bibitem{wu2020lite}
Z.~Wu, Z.~Liu, J.~Lin, Y.~Lin, and S.~Han, ``Lite transformer with long-short
  range attention,'' 2020.

\bibitem{hu2018squeeze}
J.~Hu, L.~Shen, and G.~Sun, ``Squeeze-and-excitation networks,'' in
  \emph{Proceedings of the IEEE conference on computer vision and pattern
  recognition}, 2018, pp. 7132--7141.

\bibitem{zhu2004class}
X.~Zhu and X.~Wu, ``Class noise vs. attribute noise: A quantitative study,''
  \emph{Artificial intelligence review}, vol.~22, no.~3, pp. 177--210, 2004.

\bibitem{bakshi2014intermittent}
R.~Bakhshi, S.~Kunche, and M.~Pecht, ``Intermittent failures in hardware and
  software,'' \emph{Journal of Electronic Packaging}, vol. 136, p. 011014, 03
  2014.

\bibitem{satnamintermittent}
S.~Singh, ``Decision forest for root cause analysis of intermittent faults,''
  \emph{IEEE TRANSACTIONS ON HUMAN-MACHINE SYSTEMS}, vol.~42, p. 1818, 11 2012.

\bibitem{wiki:Semiconductor_device_fabrication}
Wikipedia, ``{Semiconductor device fabrication} --- {W}ikipedia{,} the free
  encyclopedia,''
  \url{http://en.wikipedia.org/w/index.php?title=Semiconductor\%20device\%20fabrication&oldid=1037063416},
  2021, [Online; accessed 05-August-2021].

\bibitem{hochreiter1997long}
S.~Hochreiter and J.~Schmidhuber, ``Long short-term memory,'' \emph{Neural
  computation}, vol.~9, no.~8, pp. 1735--1780, 1997.

\bibitem{karim2019multivariate}
F.~Karim, S.~Majumdar, H.~Darabi, and S.~Harford, ``Multivariate lstm-fcns for
  time series classification,'' \emph{Neural Networks}, vol. 116, pp. 237--245,
  2019.

\bibitem{kingma2017adam}
D.~P. Kingma and J.~Ba, ``Adam: A method for stochastic optimization,'' 2017.

\bibitem{zhang2021autoencoder}
C.~Zhang and S.~Bom, ``Auto-encoder based model for high-dimensional imbalanced
  industrial data,'' 2021.

\bibitem{chao2021soft}
C.~Zhang, J.~Yella, Y.~Huang, X.~Qian, S.~Petrov, A.~Rzhetsky, and S.~Bom,
  ``Soft sensing transformer:hundreds of sensors are worth a single word,'' in
  \emph{2021 IEEE International Conference on Big Data (Big Data)}.\hskip 1em
  plus 0.5em minus 0.4em\relax IEEE, 2021.

\bibitem{xiaoye2021soft}
X.~Qian, C.~Zhang, J.~Yella, Y.~Huang, S.~Petrov, M.-C. Huang, and S.~Bom,
  ``Soft sensing model visualization: Fine-tuning neural network from what
  model learned,'' in \emph{2021 IEEE International Conference on Big Data (Big
  Data)}.\hskip 1em plus 0.5em minus 0.4em\relax IEEE, 2021.

\bibitem{yu2021grassnet}
Y.~Huang, C.~Zhang, J.~Yella, X.~Qian, S.~Petrov, Y.~Tang, X.~Zhu, and S.~Bom,
  ``Grassnet: Graph soft sensing neural networks,'' in \emph{2021 IEEE
  International Conference on Big Data (Big Data)}.\hskip 1em plus 0.5em minus
  0.4em\relax IEEE, 2021.

\end{thebibliography}

\vspace{12pt}

\end{document}